\pdfoutput=1
\documentclass{article}
\usepackage{nips15submit_e,times,graphicx}
\usepackage{hyperref}
\usepackage{url}
\usepackage{subfig,xspace,amsthm,multirow}
\usepackage[pointedenum]{paralist}
\usepackage{listings}
\usepackage[ruled]{algorithm2e}
\usepackage{amsmath,amssymb}

\newcommand*\samethanks[1][\value{footnote}]{\footnotemark[#1]}
\newtheorem{theorem}{Theorem}
\newtheorem{corollary}{Corollary}
\title{Distributed Training of Structured SVM}
 \author{
	 Ching-pei Lee\thanks{Most parts of this work was done when the authors were at University of Illinois.}\\
		 University of Wisconsin-Madison\\
		 \texttt{ ching-pei@cs.wisc.edu} \\
		 \And
		 Kai-Wei Chang\samethanks \\
		 Microsoft Research\\
		 \texttt{kw@kwchang.net}\\
		 \And
		Shyam Upadhyay\\
University of Illinois at Urbana-Champaign \\
\texttt{upadhya3@illinois.edu}\\
\And
		Dan Roth \\
University of Illinois at Urbana-Champaign \\
\texttt{danr@illinois.edu}
}
 \nipsfinalcopy

\def\N{{ \mathcal{N}}}

\def\bx{{\boldsymbol x}}
\def\by{{\boldsymbol y}}

\def\bv{{\boldsymbol v}}

\def\bw{{\boldsymbol w}}
\def\bv{{\boldsymbol v}}

\def\bd{{\boldsymbol d}}
\def\bv{{\boldsymbol v}}
\def\b1{{\boldsymbol 1}}

\def\Bxi{\boldsymbol \xi}
\def\AL{{\boldsymbol{\alpha}}}

\def\by{{\boldsymbol{y}}}

\def\bx{{\boldsymbol{x}}}

\def\AL{{\boldsymbol{\alpha}}}

\def\bw{{\boldsymbol{w}}}

\def\bxi{{\boldsymbol{x}_i}}
\def\byi{{\boldsymbol{y}_i}}

\def\POS{{POS}\xspace}
\def\dep{{DP}\xspace}
\def\ADMM{{\sc ADMM-Struct}\xspace}
\def\bqo{{\sc BQO-Struct}\xspace}
\def\perceptron{{\sc Distributed Perceptron}\xspace}
\def\xset{{\mathcal{X}}}
\def\yset{{\mathcal{Y}}}

\def\b0{{\bf{0}}}
\def\be{{\mathbf{e}}}

\begin{document}
\maketitle
\begin{abstract}
Training structured prediction models is time-consuming.
However, most existing approaches only use a single machine,
thus, the advantage of computing power and the capacity for
larger data sets of multiple machines have not been exploited.
In this work, we propose an efficient algorithm for
distributedly training structured support vector
machines based on a
distributed block-coordinate descent method.
Both theoretical and experimental results indicate that our
method is efficient.
\end{abstract}
\section{Introduction}
Many tasks in natural language processing and computer vision can be
formulated as structured prediction problems, where the goal is to assign
values to mutually dependent variables. The
inter-dependencies constitute the ``structure''.   To fully
exploit the rich representation of the structures, it is essential to
use large amount of data.
However, in practice,
only a limited amount of data can be used to
train a structured model because
most current approaches for structured learning are confined to a single machine,
which imposes a limit on memory and disk capacity.
For linear classification, this
problem has been addressed by distributed training
algorithms (see, e.g., \cite{YZ14a-short, CYL14a, LeeRo15,
 CZ12a-short, ACDL14}).  However, there is little work on developing distributed algorithms for general structured learning.

Moreover, most existing distributed training algorithms for linear classification
rely on certain properties of the objective function (e.g., differentiability). However,
directly applying these methods to structured learning results in inferior convergence rates.
For example, dissolve-struct\footnote{\url{http://dalab.github.io/dissolve-struct/}.} uses
the framework in \cite{MJ14a} for structured SVM, but this leads to a convergence rate that is only sublinear.

There are several challenges in distributed structured learning.
First, the features vectors, which extracted from both the input and the output structures,
are often generated on-the-fly during the training process.
Synchronizing their indices across different machines may introduce additional overhead.
Second, the training time of an learning algorithm consists of three parts: 1) communication,
2) inference, and 3) learning.
It is important to balance these three factors. This is in contrast to linear classification, where
communication is often the only bottleneck.

In this work, we address these challenges and extend the recently
proposed distributed
box-constrained quadratic optimization algorithm (BQO)~\cite{LeeRo15}
for structured support vector machines (SSVM)~\cite{TJHA05,BT04a}.
We show that the global linear convergence rate
$O(\log(1/\epsilon))$ can be obtained, even if the objective function of SSVM is non-smooth.
This result is substantial, because reducing the outer iterations saves
the time taken to solve the costly sub-problems.
Moreover,  the per-machine local sub-problems in BQO can be formed as
small SSVM problems, which can be efficiently solved by off-the-shelf
solvers. This enables us to leverage the well-studied single-machine
structured learning methods such as the dual coordinate decent
algorithm \cite{ChangYi13}.  Experiments show that our algorithm is
efficient and is therefore suitable for training large-scale
structured models.

\paragraph{Existing Works.}
 A distributed structured Perceptron algorithm using the map-reduce framework is proposed in \cite{McDonaldHaMa10}.
A structured Perceptorn algorithm with mini-batch updates is discussed in \cite{ZhaoHuang13}. However,
it is unclear how to extend their algorithm on a multi-core machine to a distributed setting.
When the inference problem is formulated as a factor graph,
\cite{LBGKH12,SHPU12} proposed to split the graph-based optimization problem into sub-problems, where each problem deals with a sub-graph.
Then each machine solves a sub-problem in parallel and communicates with each other to enforce
consistency. The convergence rate of this type of approaches is unclear.
Moreover, our approach distributes instances instead of sub-graphs and
is more suitable for problems with unfactorable structures and/or many instances (e.g., parsing, sequence tagging, and alignment). A
simple distributed implementation of cutting plane method\footnote{\url{http://alexander-schwing.de}.} is also available.
They solve the inference problems in parallel and use one machine
to learn the model. This type of approaches requires many outer iterations,
and they are empirically slow even in a single machine multi-core setting (see \cite{ChangSrRo13}).

\section{Structured Support Vector Machine}
\label{sec:ssvm}
Given a set of observations $\{(\bx_i,\by_i)\}_{i=1}^l$,
where $\bx_i \in \xset$ are instances with the corresponding annotated structure
$\byi \in \yset_i$, and $\yset_i$ is the set of all feasible
structures for $\bxi$, SSVM solves
\begin{equation}
	\label{eq:ssvm}
	\min\nolimits_{\bw, \Bxi} \quad (1/2) \bw^T\bw + C \sum\nolimits_{i=1}^l
	\ell(\xi_i) \quad
	\text{ s.t. } \quad \bw^T \phi(\by, \by_i, \bx_i)
	\geq \Delta (\by_i, \by) - \xi_i, \forall i, \forall \by \in \yset_i,
\end{equation}
where
$C>0$ is a predefined parameter.
$\phi(\by,\by_i,\bx_i) = \Phi(\bx_i, \by_i) - \Phi(\bx_i, \by),$ and
$\Phi(\bx,\by)$ is the generated feature vector depending on
both the input $\bx$ and the structure $\by$.
$\ell(\xi)$ is the loss term to be minimized,
and the loss function $\Delta(\by,\by_i)\geq 0$ is a metric that represents the distance
between structures. In this paper, we consider the L2-loss, $\ell(x) =
x^2$.\footnote{
	The dual form of L1-loss SSVM has an additional linear constraint,
	which can be viewed as a polyhedron.
	Thus the algorithm is still applicable
	and the convergence rate analysis technique
	is still valid.}

We consider solving Eq. \eqref{eq:ssvm} in its dual form.
Let $\AL$ be the
vector of the dual variables with dimension $\prod |\yset_i|$, $\otimes$ be the Kronecker product,
and $\be$ be the vector of ones,
the dual of \eqref{eq:ssvm} can be written as,
\begin{equation}
  \label{eq:dual-ssvm}
  \begin{split}
	\min\nolimits_{\AL\geq \b0}&\quad f(\AL) \equiv (1/2) \AL^T \left(Q+A/2C\right) \AL
		- \bv^T \AL, \\
	Q_{(i,\by_1), (j, \by_2)} &= \phi(\by_1,\by_i,\bx_i)^T \phi(\by_2, \by_j, \bx_j),
		\forall 1\leq i,j \leq l, \forall \by_1 \in \yset_i, \forall \by_2 \in \yset_j, \\
	A &= (I \otimes \be)^T(I \otimes \be), \qquad	v_{(i, \by)} = \Delta(\by_i, \by), \forall 1 \leq i \leq l, \forall \by \in \yset_i.
\end{split}
\end{equation}
From the KKT conditions,
the respective optimal solutions $\bw^*$ and $\AL^*$ to
eq. \eqref{eq:ssvm} and eq. \eqref{eq:dual-ssvm}
satisfy $\bw^* = \sum_{i,\by} \alpha^*_{i,\by} \phi(\by, \by_i,\bx_i).$
For the ease of computation, we maintain the relationship between $\bw$ and $\AL$ during the optimization process,
and treat $\bw$ as a temporary vector.

The key challenge of solving eq. \eqref{eq:dual-ssvm} is that for most
applications,	
the size of $\yset_i$ and thus the dimension of $\AL$  is exponentially large (with respect to the length of $\bx_i$),
so optimizing over all variables is unrealistic.
Efficient dual methods~\cite{ChangYi13}  maintain a small working set of dual
variables to be optimized such that the remaining variables are fixed to be zero.
These methods then  iteratively enlarge the working set until the
problem is well-optimized.\footnote{This approach is related to applying the cutting-plane methods
to solve the primal problem \eqref{eq:ssvm} \cite{TJHA05, JoachimsFiYu09}.}
The working set is selected using the sub-gradient of \eqref{eq:ssvm} with respect the current iterate.
Specifically, for each training instance $\bx_i$, we add the dual variable $\alpha_{i,\hat{\by}}$ 
corresponds to the structure $\hat{\by}$ into the working set, where
\begin{equation}
	\label{eq:loss-augmented-inference}
\hat{\by} = \arg\max\nolimits_{\by \in \yset_i} \quad \bw^T \phi(\by,\by_i,\bx_i) - \Delta(\by_i, \by).
\end{equation}
Once $\AL$ is updated, we update $\bw$ accordingly.
We call the step of computing
eq. \eqref{eq:loss-augmented-inference} ``inference'',
and call the part of optimizing eq. \eqref{eq:dual-ssvm} over a fixed working set ``learning''.
When training SSVM distributedly, the learning step involves communication 
across machines.  Therefore, inference and learning steps are both expensive.  
In the next section, we propose an algorithm that ensures fewer rounds of both parts.

\section{Distributed Box-Constrained Quadratic Optimization for SSVM}
\label{sec:distcd}
We split the training data into $K$ disjoint parts, and store
them in $K$ machines.
Eq. \eqref{eq:dual-ssvm} is a quadratic box-constrained optimization problem;
therefore, we apply the framework in \cite{LeeRo15}.
At each iteration, given the current $\AL$ and a 
symmetric positive definite $H$,
we solve
\begin{equation}
	\bd = \arg\min\nolimits_{\bd: \AL + \bd \geq \b0}\quad g_H(\bd) \equiv
	\nabla f(\AL)^T \bd + \frac{1}{2}\bd^T H \bd.
	\label{eq:direction}
\end{equation}
We then conduct a line search to decide a suitable step size
$\eta$ and update $\AL \leftarrow \AL + \eta \bd$.
The detailed description is in Algorithm \ref{alg:distcd}.
Here, we consider
$H \equiv \theta\bar{Q} +\frac{1}{2C}A + \lambda I,$
where $\lambda> 0$ is a small constant to ensure $H\succ 0$,
$\theta > 0$ can be tuned to decide how conservative the updates are,
and
\begin{equation*}
	\bar{Q}_{(i,\by_1), (j, \by_2)} = \begin{cases}
			0 &\text{ if $i,j$ are not in the same partition},\\
			\phi(\by_1, \by_i,\bx_i)^T \phi(\by_2, \by_j,\bx_j) &\text{ otherwise}.
		\end{cases}
              \end{equation*}
              The choice of $H$ is based on two factors: 1)
              To converge fast, $H$ should be an approximation of the real Hessian; 
              2) To solve eq. \eqref{eq:direction} without
              incurring communication cost across different machines, $H$ should be decomposable to sub-matrices, where each sub-matrix
             uses  information from data stored on one machine.
Our design of $H$ enables eq.
\eqref{eq:direction} to be split into $K$
sub-problems and solved locally.
Each sub-problem can be rewritten as a
SSVM dual problem.
Thus, one can adopt any single-machine SSVM solver
(e.g., \cite{JoachimsFiYu09, TJHA05, CHT09a, ChangYi13,SPSS11})
to solve it.
After \eqref{eq:direction} is solved,
we compute
\begin{equation}
	\Delta \bw \equiv \sum\nolimits_{i,\by} \bd_{i,\by} \phi(\by, \by_i,\bx_i)
	\label{eq:delta}
\end{equation}
by an {\em allreduce} operation that communicates information between machines.
This information also synchronizes the model for conducting inferences to enlarge the working set.
Using $\Delta \bw$, an exact line search for deciding the optimal step size $\eta^*$ can be conducted.
\begin{equation*}
	\frac{\partial f(\AL + \eta \bd)}{\partial \eta} = 0 \Rightarrow
	\eta^* = \frac{-\nabla f(\AL)^T\bd}{\bd^T(Q+A/2C)\bd} =
	-\frac{\bw^T \Delta \bw + \AL^T(A/2C)\bd - \bv^T \bd}{\Delta \bw^T \Delta \bw + \bd^T(A/2C)\bd}.
\end{equation*}
To ensure feasibility, we take the final step size $\eta$ to be
\begin{equation}	
\eta = \min(\max\{\eta'\mid \AL + \eta' \bd \geq \b0\}, \eta^*).
\label{eq:eta}
\end{equation}
\begin{algorithm}[t]
  \caption{A box-constrained quadratic optimization algorithm for solving \eqref{eq:ssvm}}
	\label{alg:distcd}
	\begin{compactenum}
		\item $\bw \leftarrow \b0, \AL \leftarrow \b0$.
		\item  For $t=0,1,\ldots$ \quad (outer iteration)
		\begin{compactenum}
		\item Use the current $\bw$ to
		solve \eqref{eq:direction} to get $\bd$ distributedly in
			$K$ machines.
		\item Use {\em allreduce} to obtain $\Delta \bw$ in
                  eq. \eqref{eq:delta}.
		\item Compute $\eta$ by eq. \eqref{eq:eta} with another $O(1)$ communication.
		\item $\AL \leftarrow \AL + \eta \bd$; $\bw \leftarrow \bw + \eta \Delta \bw$.
		\end{compactenum}
	\end{compactenum}
\end{algorithm}

Following the analysis in \cite{LeeRo15},
we can show the following convergence result for
Algorithms \ref{alg:distcd}.
\begin{theorem}
	\label{thm:distcd}
	Algorithm \ref{alg:distcd} has global linear convergence
	when the exact solution of \eqref{eq:direction} is obtained at each iteration and $H\succ 0$.
      \end{theorem}
In practice,
obtaining the exact solution of \eqref{eq:direction} is time-consuming.
We show that global linear convergence still holds when \eqref{eq:direction} is solved approximately.
\begin{corollary}
        Let $\bd^*$ be the optimal solution of \eqref{eq:direction}.
        If for some constant $\gamma \in [0,1)$ and for all $t$,
        the update direction $\bd$  satisfies
        $\gamma |g_H(\bd^*)| \leq |g_H(\bd)|$ with $H \succ 0$,
      	then Algorithm \ref{alg:distcd} converges with a global linear rate.
\end{corollary}
Since $\gamma$ is arbitrary, for any sub-problem solver that strictly decreases the function value, we can easily obtain a value of $\gamma < 1$.

The communication step in eq. \eqref{eq:delta}
requires machines to communicate a vector of $O(n)$.
The actual cost of this communication depends on the network setting and usually grows with $K$.
We note that solving \eqref{eq:direction} approximately results in more iterations and thus more rounds of communication,
but requires fewer inference calls.
Thus this is a trade-off between communication and inference.
For many applications, inference is much more expensive than communication,
thus the balance between these two factors is worth studying empirically.

\paragraph{Model Consistency}
Unlike binary classification, while learning a structured model, features are usually generated on-the-fly because 
the feature set depends on the structures the solver has seen so far.
If each machine maintains its own feature mapping, the feature indices will be 
inconsistent across machines.
One potential solution is to synchronize the feature mappings
at each round. However, this approach incurs a huge communication overhead.
To tackle this issue, we adapt a feature hashing strategy in \cite{WDLSA09}.
We map the features into integer values in $[0,2^d), d\in \N$ by a 
unique hashing function and use them as new feature indices, such that the size of the weight vector is at most $2^d$.
The input to this hashing function can be any object, such as an integer or a string.   
This strategy has been used in distributed environments \cite{ACDL14,LBGKH12}
for dimension reduction and fast look-up.
Here, as argued before, this techniques is crucial and efficient for distributed structured learning.


\section{Experiments}
\label{sec:exp}
We perform experiments on part-of-speech
tagging (\POS) and dependency parsing (\dep).
%
%
%
For both tasks, we use the Wall Street Journal portion 
of the Penn Treebank \cite{penn-tree-bank} with the standard split for 
training (section 02-21) and test (section 23).
For both tasks, we set $C=0.1$ for SSVM and compare the following algorithms 
using eight nodes in a local cluster.
\begin{compactenum}
	\item \bqo: the algorithm we proposed in Section \ref{sec:distcd}. We set $\theta$ to be $K$.
	\item \ADMM: the alternating directions method of multiplier \cite{SB11b}.
	\item \perceptron: a parallel structured Perceptron algorithm described in \cite{McDonaldHaMa10}.
	\item Simple average: Each machine trains a separate 
		model using the local data. The final model is obtained by averaging
		all local models.
\end{compactenum}


\begin{figure}[t]
	\centering
	\begin{tabular}{@{}c@{}c@{}c@{}c@{}}

		\subfloat[\POS]{\label{fig:pos}\includegraphics[width=.25\textwidth]{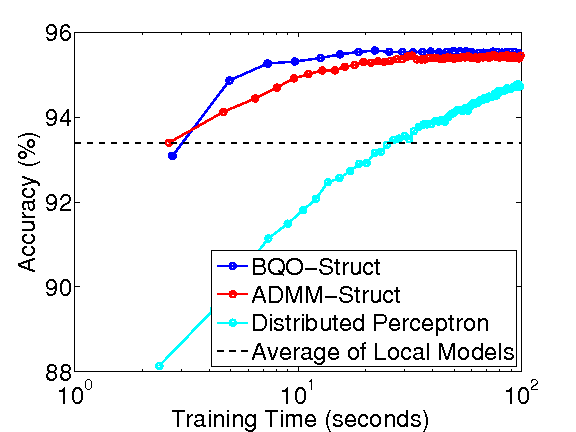}}&
		\subfloat[\dep]{\label{fig:dep}\includegraphics[width=.25\textwidth]{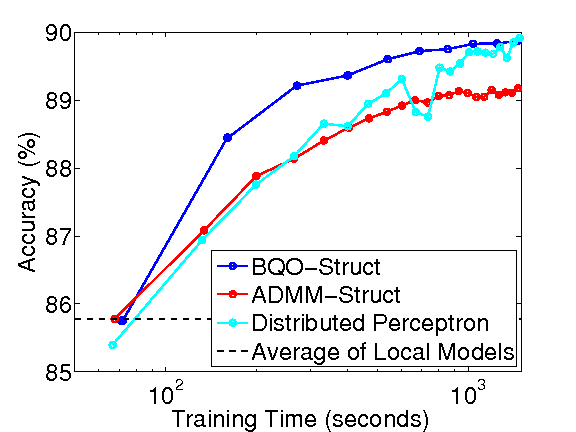}}&
		\subfloat[\POS]{\label{fig:pos-speedup}\includegraphics[width=.25\textwidth]{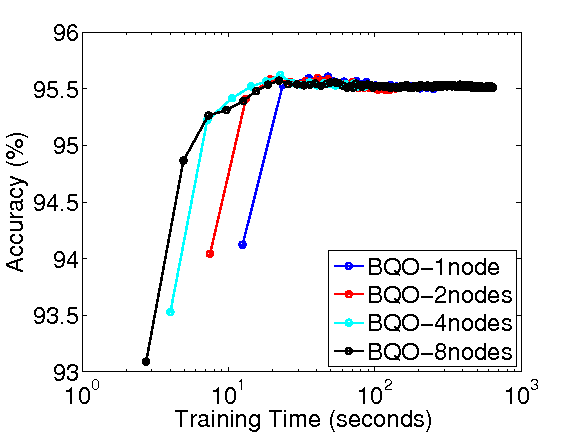}}&
		\subfloat[\dep]{\label{fig:dep-speedup}\includegraphics[width=.25\textwidth]{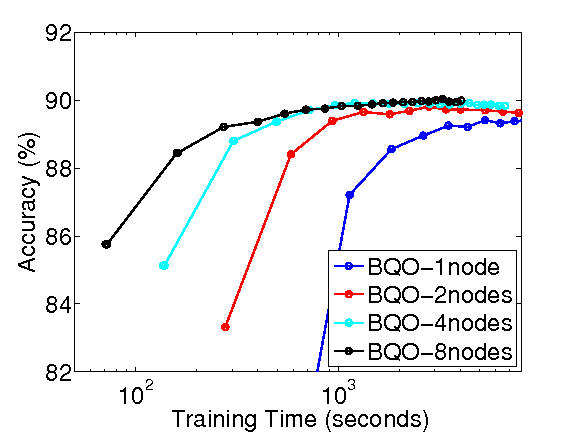}
	}
\end{tabular}
\caption{\ref{fig:pos} and \ref{fig:dep}: Comparison between different algorithms using eight nodes.
\ref{fig:pos-speedup} and \ref{fig:dep-speedup}: Performance of \bqo using different number of machines.
Training time is in {\it log scale}.}
	\label{fig:exp}
\end{figure}


The sub-problems in \ADMM and \bqo are solved by the dual coordinate descent solver proposed in
\cite{ChangYi13}, which is shown to be empirically faster than
other existing methods.
To have a fair comparison,
we use the same setting for solving sub-problems when possible.

Because different methods solve different objectives,
we compare the test performance along training time.
Figure \ref{fig:exp} shows the results.
\bqo performs the best in both tasks,
confirming its fast theoretical convergence rate.
We further investigate the speedup of \bqo
in Figures \ref{fig:pos-speedup}-\ref{fig:dep-speedup}.
This also serves as a comparison between our distributed algorithm and the state-of-the-art single-machine SSVM solver.
For the time-consuming task \dep,
the speedup is significant because a large portion of the training time is spent on inference.
Parallelizing this part can achieve nearly linear speedup.
While for \POS, because the training time using a single machine is already fast enough,
using multiple machines does not improve the training time much.

Overall, this work addresses the challenge of training structured SVM problems
in a distributed setting
and
proposes an algorithm with fast convergence rate and
good empirical performance.
We hope this work will inspire
more applications of structured learning with large volume of training data
to improve the performance on structured learning tasks.
{
\scriptsize
This research was supported by the Multimodal Information Access \& Synthesis Center at UIUC, part of CCICADA, a DHS Science and Technology Center of Excellence and by DARPA under agreement number FA8750-13-2-0008. The U.S. Government is authorized to reproduce and distribute reprints for Governmental purposes notwithstanding any copyright notation thereon. The views and conclusions contained herein are those of the authors and should not be interpreted as necessarily representing the official policies or endorsements, either expressed or implied, of DARPA or the U. S. Government.
}

\bibliography{cj,cited-compact,ccg-compact,my}

\begin{thebibliography}{10}

\bibitem{ACDL14}
A.~Agarwal, O.~Chapelle, M.~Dud\'{i}k, and J.~Langford.
\newblock A reliable effective terascale linear learning system.
\newblock {\em Journal of Machine Learning Research}, 2014.

\bibitem{SB11b}
S.~Boyd, N.~Parikh, E.~Chu, B.~Peleato, and J.~Eckstein.
\newblock Distributed optimization and statistical learning via the alternating
  direction method of multipliers.
\newblock {\em Foundations and Trends in Machine Learning}, 3(1):1--122, 2011.

\bibitem{ChangSrRo13}
K.-W. Chang, V.~Srikumar, and D.~Roth.
\newblock Multi-core structural {SVM} training.
\newblock In {\em ECML}, 2013.

\bibitem{ChangYi13}
M.-W. Chang and W.-T. Yih.
\newblock Dual coordinate descent algorithms for efficient large margin
  structural learning.
\newblock {\em Transactions of the Association for Computational Linguistics},
  2013.

\bibitem{MJ14a}
M.~Jaggi, V.~Smith, M.~Tak{\'a}{\v{c}}, J.~Terhorst, T.~Hofmann, and M.~I.
  Jordan.
\newblock Communication-efficient distributed dual coordinate ascent.
\newblock In {\em Advances in Neural Information Processing Systems 27}. 2014.

\bibitem{JoachimsFiYu09}
T.~Joachims, T.~Finley, and C.-N. Yu.
\newblock Cutting-plane training of structural {SVM}s.
\newblock {\em Machine Learning}, 2009.

\bibitem{LeeRo15}
C.-P. Lee and D.~Roth.
\newblock Distributed box-constrained quadratic optimization for dual linear
  {SVM}.
\newblock In {\em ICML}, 2015.

\bibitem{CYL14a}
C.-Y. Lin, C.-H. Tsai, C.-P. Lee, and C.-J. Lin.
\newblock Large-scale logistic regression and linear support vector machines
  using {S}park.
\newblock In {\em Proceedings of the IEEE International Conference on Big
  Data}, pages 519--528, 2014.

\bibitem{LBGKH12}
Y.~Low, D.~Bickson, J.~Gonzalez, C.~Guestrin, A.~Kyrola, and J.~M. Hellerstein.
\newblock Distributed graphlab: A framework for machine learning and data
  mining in the cloud.
\newblock {\em Proceedings of the VLDB Endowment}, 5(8), 2012.

\bibitem{penn-tree-bank}
M.~P. Marcus, B.~Santorini, and M.~A. Marcinkiewicz.
\newblock Building a large annotated corpus of {E}nglish: The {P}enn
  {T}reebank.
\newblock {\em Computational Linguistics}.

\bibitem{McDonaldHaMa10}
R.~McDonald, K.~Hall, and G.~Mann.
\newblock Distributed training strategies for the structured {Perceptron}.
\newblock In {\em ACL}, 2010.

\bibitem{SHPU12}
A.~G. Schwing, T.~Hazan, M.~Pollefeys, and R.~Urtasun.
\newblock Efficient structured prediction with latent variables for general
  graphical models.
\newblock In {\em ICML}, 2012.

\bibitem{SPSS11}
S.~K. Shevade, B.~P., S.~Sundararajan, and S.~S. Keerthi.
\newblock A sequential dual method for structural {SVM}s.
\newblock In {\em SDM}, 2011.

\bibitem{BT04a}
B.~Taskar, C.~Guestrin, and D.~Koller.
\newblock Max-margin markov networks.
\newblock In {\em Advances in Neural Information Processing Systems 16}. 2004.

\bibitem{CHT09a}
C.~H. Teo, S.~Vishwanathan, A.~Smola, and Q.~V. Le.
\newblock Bundle methods for regularized risk minimization.
\newblock {\em Journal of Machine Learning Research}, 2010.

\bibitem{TJHA05}
I.~Tsochantaridis, T.~Joachims, T.~Hofmann, and Y.~Altun.
\newblock Large margin methods for structured and interdependent output
  variables.
\newblock {\em Journal of Machine Learning Research}, 2005.

\bibitem{WDLSA09}
K.~Weinberger, A.~Dasgupta, J.~Langford, A.~Smola, and J.~Attenberg.
\newblock Feature hashing for large scale multitask learning.
\newblock In {\em ICML}, 2009.

\bibitem{CZ12a-short}
C.~Zhang, H.~Lee, and K.~G. Shin.
\newblock Efficient distributed linear classification algorithms via the
  alternating direction method of multipliers.
\newblock In {\em AISTATS}, 2012.

\bibitem{ZhaoHuang13}
K.~Zhao and L.~Huang.
\newblock Minibatch and parallelization for online large margin structured
  learning.
\newblock In {\em NAACL}, pages 370--379, 2013.

\bibitem{YZ14a-short}
Y.~Zhuang, W.-S. Chin, Y.-C. Juan, and C.-J. Lin.
\newblock Distributed {N}ewton method for regularized logistic regression.
\newblock In {\em PAKDD}, 2015.

\end{thebibliography}
\bibliographystyle{abbrv}
\end{document}